# Rock mechanics modeling based on soft granulation theory

H. Owladeghaffari
*Dept .Mining &Metallurgical Engineering, Amirkabir University of Technology, Tehran, Iran*

**ABSTRACT:** This paper describes application of information granulation theory, on the design of rock engineering flowcharts. Firstly, an overall flowchart, based on information granulation theory has been highlighted. Information granulation theory, in crisp (non-fuzzy) or fuzzy format, can take into account engineering experiences (especially in fuzzy shape-incomplete information or superfluous), or engineering judgments, in each step of designing procedure, while the suitable instruments modeling are employed. In this manner and to extension of soft modeling instruments, using three combinations of Self Organizing Map (SOM), Neuro-Fuzzy Inference System (NFIS), and Rough Set Theory (RST) crisp and fuzzy granules, from monitored data sets are obtained. The main underlined core of our algorithms are balancing of crisp(rough or non-fuzzy) granules and sub fuzzy granules, within non fuzzy information (initial granulation) upon the "open-close iterations". Using different criteria on balancing best granules (information pockets), are obtained. Validations of our proposed methods, on the data set of in-situ permeability in rock masses in Shivashan dam, Iran have been highlighted.

## 1. INTROUDCTION

In the recent years, developing new indirect analysis methods has opened new horizons in rock engineering solutions. Tendency to the micro-view of the natural events in the rock based systems, upon the high speed PC technology; behind large-scale investigations, has allocated new challenges in this track. Transition from a general (overall) view in to the detailed descriptions, can be interpreted as relations (inter-extra relations) of the commuted "information packages", got from accumulations of data, experience, novelty and other effective agents. Such construction of the "whole" from "part" (granules) is the current behavior of human cognition. Among the basic concepts which underlie human cognition there are three remarkable sides, which are: granulation, organization, and causation. Granulation involves decomposition of whole into parts; organization involves integration of parts in to whole; and causation relates to association of causes with effects [1].

Under this view from the discritization (meshing, blocking, latticing…) of the interior or boundary of a field to the solving steps (thinking) of the problem are the perspectives of granulation. We called first level of granulation as "hard granulation", and second level as "soft granulation". To better understand of the meaning of hard and soft granulation, we reproduce the general rock engineering design flowchart in figure1 [2]. Level1 can be supposed as a hard granulation where level2 is related with soft granulation.

Clearly, in soft granulation; we are approaching to the real human cognition, whereas in hard packing the machine computations are distinguished. Let us, consider the last class in level2 (in category D): internet based system. Interestingly, this category shows how the discriminated projects, under the virtual world, employ the distributed information granules. Plainly, the contributions of any projects and the sub-sets of granules in construction of this general network are affected from the several parameters, concluded in "granulation level" factor.
In this paper, we interest to tack in to account soft granulation in rock system. Upon this, by focusing in two categories C-1 and C-2, in figure1, we develop different soft granulation methods based on intelligent systems and approximate reasoning methods. Added to this, the bridging between hard and soft granulation is abstracted.

The most main distinguished facets of the soft granules are: set theory, interval analysis, fuzzy set, rough set. Each of these theories considers part of uncertainty of information (data, words, pictures...). Due to association of uncertainty and vagueness with the monitored data set, particularly, resulted from the in-situ tests (such lugeon test), accounting relevant approaches such probability, Fuzzy Set Theory (FST) and Rough Set Theory (RST) to knowledge acquisition, extraction of rules and prediction of unknown cases, more than the past have been distinguished. Zadeh has emphasized the role of FST in geosciences will be increased during future years [3].

The RST introduced by Pawlak has often proved to be an excellent mathematical tool for the analysis of a vague description of object [4], [5]. The adjective vague, referring to the quality of information, means inconsistency, or ambiguity which follows from information granulation. The rough set philosophy is based on the assumption that with every object of the universe, is associated a certain amount of information, expressed by means of some attributes used for object description. The indiscernibility relation (similarity), which is a mathematical basis of the rough set theory, induces a partition of the universe in to blocks of indiscernible objects, called elementary sets, which can be used to build knowledge about a real or abstract world. Precise condition rules can be extracted from a discernibility matrix. Application of RST in different fields of the applied sciences has been reported [6], [7], but developing of such system (based on approximate analysis) in rock engineering have not been outstanding, relatively. Figure 2 shows a general procedure, in which the IGT accompanies by a predefined project based rock engineering design. After determination of constraints and the associated rock engineering considerations, the initial granulation of information as well as numerical (data base) or in linguistic formats is accomplished. Improvement of modeling instruments based upon IGs, whether in independent or affiliated shape with hard computing methods (such fuzzy finite element, fuzzy boundary element, stochastic finite element…) are new challenges in the current discussion. In this study, under "modeling instruments" box, we propose three algorithms; namely successive elicitation of crisp (non-fuzzy), fuzzy and rough granulations: Self Organizing Neuro-Fuzzy Inference System (Random and Regular neuron growth), in an abbreviated manner: SONFIS-R, SONFIS-AR; and Self Organizing Rough Set Theory (SORST).

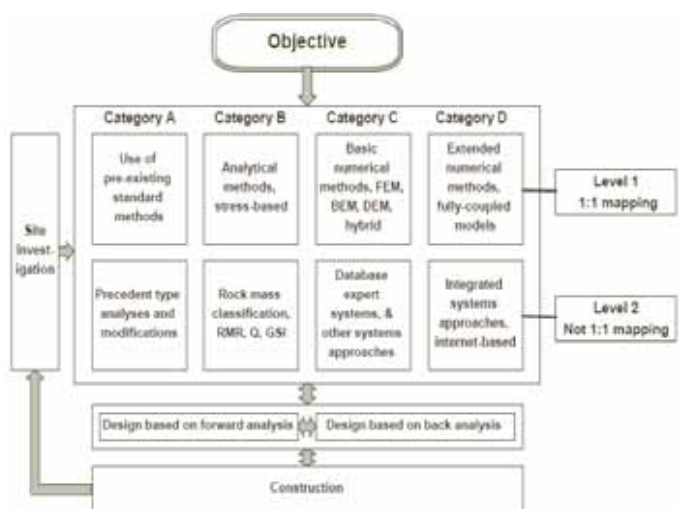

Fig.1. one of the last general flowcharts to rock engineering design [2]

In figure 3, we have concluded a summary of current overall granulation in a rock project that leads to the formation of fuzzy granules on the attributes (properties) of joints. Figure4 show how one usually employs granulation procedure to permeability analysis in a dam site, instinctively.

The rest of paper has been organized as section 2: preliminaries on some soft granulation methods, i.e. SOM, NFIS, and RST in next section, we propose three main algorithms and part 4 covers a practical instance, describes how the soft granules ensue a relatively complete analysis on the permeability of Shivashan dam site, in Iran.

## 2. PRELIMINARIES

### 2.1. Self Organizing feature Map (SOM)

Kohonen's SOM algorithm has been well renowned as an ideal candidate for classifying input data in an unsupervised learning way [8]. Kohonen self-organizing networks (Kohonen feature maps or topology-preserving maps) are competition-based network paradigm for data clustering. The learning procedure of Kohonen feature maps is similar to the competitive learning networks. The main idea behind competitive learning is simple; the winner takes all. The competitive transfer function returns neural outputs of 0 for all neurons except for the winner which receives the highest net input with output 1.

SOM changes all weight vectors of neurons in the near vicinity of the winner neuron towards the input vector. Due to this property SOM, are used to reduce the dimensionality of complex data (data clustering). Competitive layers will automatically learn to classify input vectors, the classes that the competitive layer finds are depend only on the distances between input vectors [8].

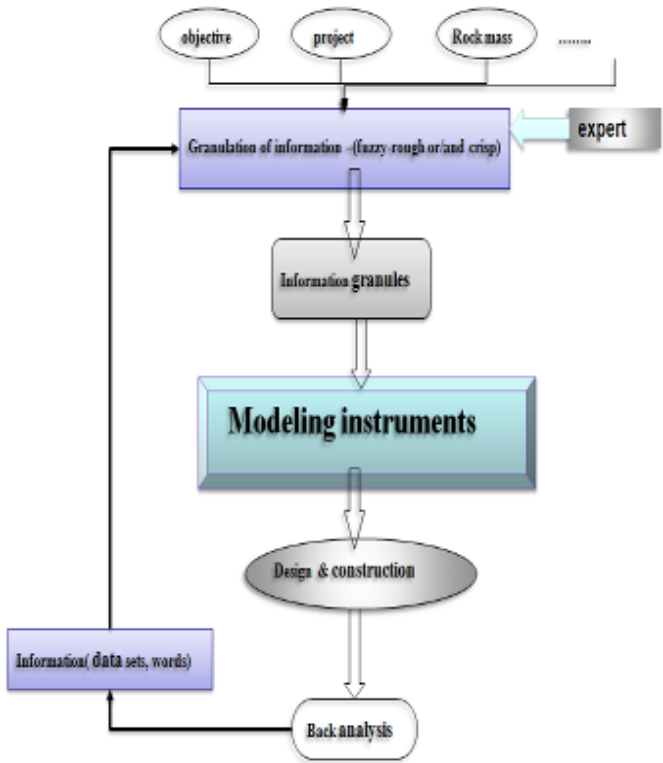

Fig.2. A general methodology for back analysis based on IGT

### 2.2. Neuro-fuzzy inference system ( NFIS)

There are different solutions of fuzzy inference systems. Two well-known fuzzy modeling methods are the Tsukamoto fuzzy model and Takagi–Sugeno–Kang (TSK) model. In the present work, only the TSK model has been considered.

The TSK fuzzy inference systems can be easily implanted in the form of a so called Neuro-fuzzy network structure .in this study, we have employed an adaptive neuro-fuzzy inference system [9].

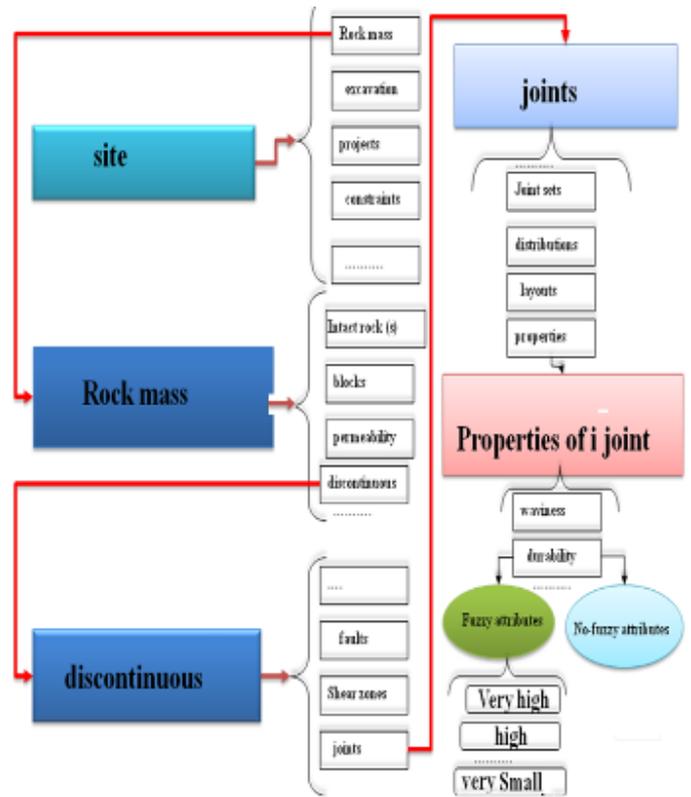

Fig.3. Current overall granulation in a rock project

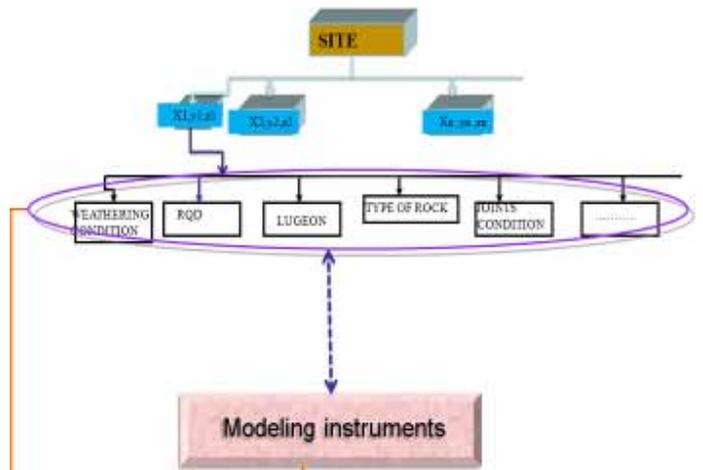

Fig.4. Granulation procedure to permeability analysis in a dam site

One of the most important stages of the Neuro-fuzzy TSK network generation is the establishment of the inference rules. Often used is the so-called grid method, in which the rules are defined as the combinations of the membership functions for each input variable. If we split the input variable range into a limited number (say $n_i$ for $i=1, 2... n$) of membership functions, the combinations of them lead to many different inference rules. The problem is that these combinations correspond in many cases to the regions of no data, and hence a lot of them may be deleted. This problem can be solved by using the fuzzy self-organization algorithm. This algorithm splits the data space into a specified number of overlapping clusters. Each cluster may be associated with the specific rule of the center corresponding to the center of the appropriate cluster. In this way all rules correspond to the regions of the space-containing majority of data and the problem of the empty rules can be avoided. The ultimate goal of data clustering is to partition the data into similar subgroups. This is accomplished by employing some similar measures (e.g., the Euclidean distance) [9].

In this paper data clustering is used to derive membership functions from measured data, which, in turn, determine the number of If-Then rules in the model (i.e., rules indication).

The method employed in this paper is the subtractive clustering method, proposed by Yager as one of the simplest clustering methods [10].

### 2.3. Rough Set Theory (RST)

The rough set theory introduced by Pawlak [4], [5] has often proved to be an excellent mathematical tool for the analysis of a vague description of object. The adjective vague referring to the quality of information means inconsistency, or ambiguity which follows from information granulation.

An information system is a pair $S=<U, A>$, where $U$ is a nonempty finite set called the universe and $A$ is a nonempty finite set of attributes. An attribute a can be regarded as a function from the domain $U$ to some value set $V_a$. An information system can be represented as an attribute-value table, in which rows are labeled by objects of the universe and columns by attributes. With every subset of attributes $B \subseteq A$, one can easily associate an equivalence relation $I_B$ on $U$:

$$I_B = \{(x,y) \in U : for\ every\ a \in B, a(x) = a(y)\} \quad (1)$$

Then, $I_B = \bigcap_{a \in B} I_a$.

If $X \subseteq U$, the sets $\{x \in U : [x]_B \subseteq X\}$ and $\{x \in U : [x]_B \cap X \neq \varphi\}$, where $[x]_B$ denotes the equivalence class of the object $x \in U$ relative to $I_B$, are called the B-lower and the B-upper approximation of X in S and denoted by $\underline{BX}$ and $\overline{BX}$, respectively. Consider $U = \{x_1, x_2, ..., x_n\}$ and $A = \{a_1, a_2, ..., a_n\}$ in the information system $S = \prec U, A \succ$.

By the discernibility matrix $M(S)$ of $S$ is meant an $n*n$ matrix such that

$$c_{ij} = \{a \in A : a(x_i) \neq a(x_j)\} \quad (2)$$

A discernibilty function $f_s$ is a function of m Boolean variables $a_1...a_m$ corresponding to attribute $a_1...a_m$, respectively, and defined as follows:

$$f_s(a_1,...,a_m) = \wedge \{\vee(c_{ij}) : i, j \leq n, j \prec i, c_{ij} \neq \varphi\} \quad (3)$$

Where $\vee(c_{ij})$ is the disjunction of all variables with $a \in c_{ij}$. Using such discriminant matrix the appropriate rules are elicited.

One of the main parameters in the covering of the obtained rule is "dependency rule or strength ". Let; we have a rule in the disjunctive normal form (D.N.F), for instance:

$$\underbrace{(a_1 \wedge ...a_n) \vee (b_1 \wedge ... \wedge a_m).....}_{P_i} \to d_i\ (decision\ attribute)$$

The dependency factor $df_i$ is given by:

$$df_i = \frac{Card(POS_{B_i}(d_i))}{Card(U_i)} \quad (4)$$

Where $POS_{B_i}(d_i) = \bigcup_{X \in I_{d_i}} \underline{B}_i(X)$, and $\underline{B}_i(X)$ is the lower approximation of $X$ with respect to $\underline{B}_i(X)$. $\underline{B}_i$ is the set of condition attributes( say inputs) occurring in the rule. $POS_{B_i}(d_i)$ is the positive region of the decision class $d_i$ that can be surely described by attributes $\underline{B}_i$ [6]. The existing induction algorithms use one of the following strategies:

(a) Generation of a minimal set of rules covering all objects from a decision table;
(b) Generation of an exhaustive set of rules consisting of all possible rules for a decision table;
(c) Generation of a set of `strong' decision rules, even partly discriminant, covering relatively many objects each but not necessarily all objects from the decision table [11]. In this study we have developed RST in MatLab7, and on this added toolbox other appropriate algorithms have been prepared.

### 3. PROPOSED ALGORITHMS

In the whole of our algorithms, we use four basic axioms upon the balancing of the successive granules:

*Step (1): dividing the monitored data into groups of training and testing data*

*Step (2): first granulation (crisp) by SOM or other crisp granulation methods*

   *Step (2-1): selecting the level of granularity*

*randomly or depend on the obtained error from the NFIS or RST (regular neuron growth)*

*Step (2-2): construction of the granules (crisp).*
*Step (3): second granulation (fuzzy or rough IGs) by NFIS or RST*

*Step (3-1): crisp granules as a new data.*
*Step (3-2): selecting the level of granularity; (Error level, number of rules, strength threshold...)*
*Step (3-3): checking the suitability. (Close-open iteration: referring to the real data and reinspect closed world)*
*Step (3-4): construction of fuzzy/rough granules.*
Step (4): extraction of knowledge rules

Selection of initial crisp granules can be supposed as "Close World Assumption (CWA)". But in many applications, the assumption of complete information is not feasible, and only cannot be used. In such cases, an "Open World Assumption (OWA)', where information not known by an agent is assumed to be unknown, is often accepted [12].

Balancing assumption is satisfied by the close-open iterations: this process is a guideline to balancing of crisp and sub fuzzy/rough granules by some random/regular selection of initial granules or other optimal structures and increment of supporting rules (fuzzy partitions or increasing of lower /upper approximations ), gradually.

The overall schematic of Self Organizing Neuro-Fuzzy Inference System -Random and Regular neuron growth-: SONFIS-R, SONFIS-AR; has been shown in figure5.

In first regular granulation, we use a linear relation is given by:

$$N_{t+1} = \alpha N_t + \Delta_t ; \Delta_t = \beta E_t + \gamma \qquad (5)$$

Where $N_t = n_1 \times n_2 ; |n_1 - n_2| = Min.$ is number of neurons in SOM; $E_t$ is the obtained error (measured error) from second granulation on the test data and coefficients must be determined, depend on the used data set. Obviously, one can employ like manipulation in the rule (second granulation) generation part, i.e., number of rules.

Determination of granulation level is controlled with three main parameters: range of neuron growth, number of rules and error level. The main benefit of this algorithm is to looking for best structure and rules for two known intelligent system, while in independent situations each of them has some appropriate problems such: finding of spurious patterns for the large data sets, extra-time training of NFIS or SOM.

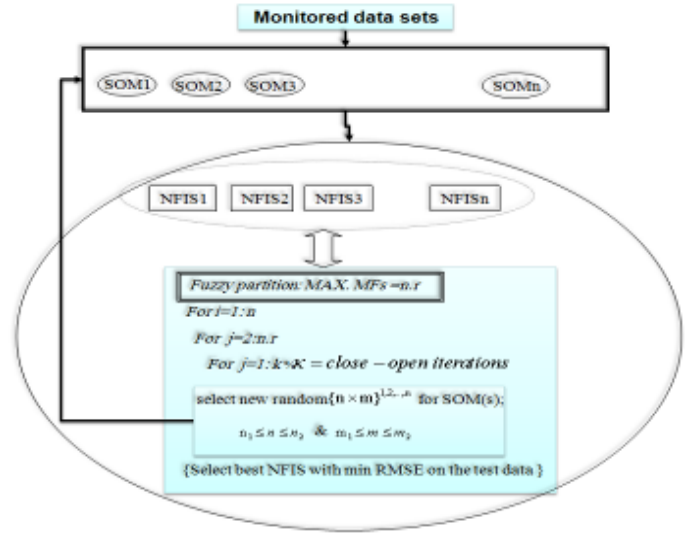

Fig.5. Self Organizing Neuro-Fuzzy Inference System (SONFIS)

In second algorithm, apart from employing hard computing methods (hard granules), RST instead of NFIS has been proposed (figure 6). Applying of SOM as a preprocessing step and discretization tool is second process. Categorization of attributes (inputs/outputs) is transferring of the attribute space to the symbolic appropriate attributes. In fact for continuous valued attributes, the feature space needs to be discretized for defining indiscernibilty relations and equivalence classes. We discretize each feature in to some levels by SOM, for example "low, medium, and high" for attribute "a".

Finer discretization may lead to better accuracy to recognizing of test data but imposes the higher cost of a computational load. However, to look for best scaled condition and decision attributes, we have developed other SORST system, upon the adaptive scaling of attributes (SORST-AS), gradually [13]. Because of the generated rules by a rough set are coarse and therefore need to be fine-tuned, here, we have used the preprocessing step on data set to crisp granulation by SOM (close world assumption).

In fact, with referring to the instinct of the human, we understand that human being want to states the events in the best simple words, sentences, rules, functions and so forth. Undoubtedly, such granules while satisfies the mentioned axiom that describe the distinguished initial structure(s) of events or immature data sets. Second SOM, as well as close world assumption, gets such dominant structures on the real data. In other word, condensation of real world and concentration on this space is associated with approximate analysis, such rough or fuzzy facets.

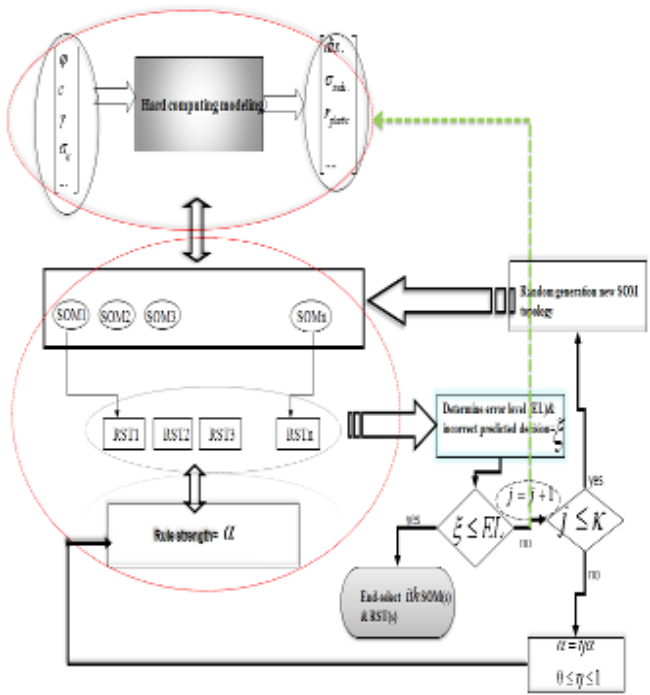

Fig.6. Bridging of hard computations and Self Organizing Rough Set Theory-Random neuron growth & adaptive strength factor (SORST-R)

Before balancing step between SOM and RST, we use a checkpoint based on granulation level, or possible best emerged granules, is assessed by setting of threshold dependency factor. In figure 5, we haven't given up the real world (as here hard granules part) and dashed arrow keeps on linkage with upper level if and only if SORST-R couldn't gratify the pre-defined constraints. Next part of this study show how these algorithms complete not1-1 mapping levels.

4. PRACTICAL EXAMPLE

In this part of our study, we pursue a practical example, which covers a comprehensive data set from lugeon test in Shivashan dam.

*4.1. Permeability assessment in Shivashan dam site-Iran*

Shivashan hydroelectric earth dam is located 45km north of Sardasht city in northwestern of Iran. Geological investigation for the site selection of the Shivashan hydroelectric power plant was made within an area of about 3 square kilometer. The width of the V-shaped valley with similarly sloping flanks, at the elevation of *1185m and 1310m* with respect to sea level are *38m* and *467m*, respectively.

At the site area, the rock type indicates generally signs of metamorphisms, which are formed by low temperature metamorphic rocks, overlain by quaternary alluvial deposits. The bedrocks consist of two types of low temperature metamorphic rocks, namely, slates and phyllites. Indeed, slate is the breadline between metamorphic and sedimentary rocks also known as a weakly metamorphosed rock. Figure 7 shows overall view of dam site. Totally, 20 boreholes have been drilled and consequently about 789 objects were resulted. Water Pressure Test (*WPT*) has used for determination of this area's permeability. *WPT* is an effective method for widely determination of rock mass permeability.

The Lugeon unit is not stated as a ratio of permeability, but to get a sense of proportion, it might be related such that: *1Lugeon=1.3*10-5 cm/s*. In practice, usually, the Lugeon test is utilized before grouting to determine quantitatively the volume of water take per unit of time [13].

A general pattern from five chief attributes of the boreholes, namely: Z=elevation of the test, L: length of the tested section, RQD and Type of the Weathering Rock Type (*T.W.R*) - see table 1- has been shown in figure 3.

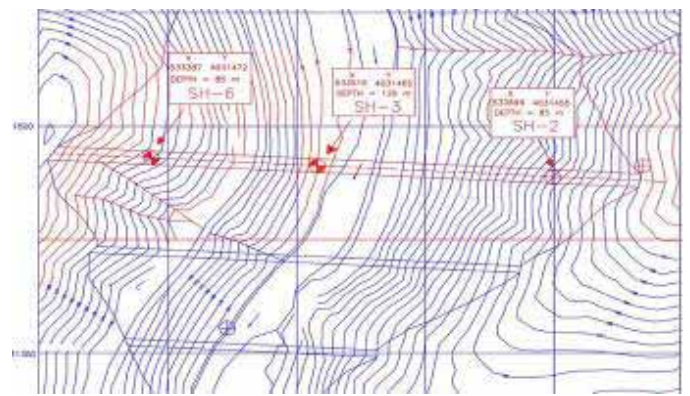

Fig. 7. Overall view of axial position at Shivashan dam

To evaluate the permeability due to the lugeon values we follow two situations: 1) utilizing of SONFIS and SORST on the five chief attributes(figure8); 2) direct application of RST and NFIS on the local coordinates of dam site (as conditional attributes) and lugeon values (as decision part) to depict 3D Iso-surfaces of lugeon variations diagrams.

Analysis of first situation is started off by setting number of close-open iteration and maximum number of rules equal to 10 and 4 in SONFIS-R, respectively. The error measure criterion in

SONFIS is Root Mean Square Error (RMSE), given as below:

$$RMSE = \sqrt{\frac{\sum_{i=1}^{m}(t_i - t_i^*)^2}{m}}$$

; Where $t_i$ is output of SONFIS and $t_i^*$ is real answer; $m$ is the number of test data (test objects). In the rest of paper, let $m=93$ and *number of training data set =600*. Figures 9, 10 indicate the results of the aforesaid system (so, performance of selected SONFIS-R on the test data). Two indicated positions in figure 10 state minimum RMSE over same rules, are near to each other. In such case, we have two opportunities: selection of SONFIS with min number of rules (*n.r*) or involved min objects. Figure 10 show second occasion. With augmenting of close-open iterations SONFIS-R or range of *n.r*, our system emerges more near min RMSE, not defiantly lowest value (figure 10-a, c).

By employing of (5) in SONFIS-AR, and $\alpha=1.01$; $\beta=.001$ and $\gamma=.5$; the general pattern of RMSE vs. neuron growth (in first layer of algorithm) can be observed (figure11). With neglecting of some details, the same trend of error fluctuation is distinguishable (highlighted by colored windows). It is worth noting that by $\alpha=.8$ and *n.r=2*; SONFIS-AR reveals a general chaos form (figure 12). The main reason of this can be followed in the first layer property: regulation of neurons in SOM may get in to the "dead station" and random selection of weights in such layer. Other reason is about the range of error vacillation. In fact in this case our system has a high senility to the error, and then to the neuron growth.

In this case, we can determine two new balance measures: durability of neurons and distribution of points in a neuron-error space. First measure gets lesser than 20 neurons while in second measure system after 100 iterations falls in the "balance hole" with nearly 50 neurons size.

Figure 13 shows our mean about structure detection. Under SONFIS-R (figure10-c) figure 13-a, declares three major clusters in *lugeon-T.W.R*. With more neurons in SOM, we can acquire like patterns but may lose the supposed balancing criteria. This proves the balance measure even if get min RMSE but losses the data distributions or major structures.

In employing of part of second algorithm (figure6), we use- -for in this case- only exact rules i.e., one decision class in right hand of an *if-then* rule. Figure 14 and 15 depict the scaling process by 1-D SOM (3 neurons) and the performance of SORST-R over 7 random selection of SOM structure, respectively. The applied error measure is:

$$MSE = \frac{\sum_{i=1}^{m}(d_i^{real} - d_i^{classified})^2}{m};$$

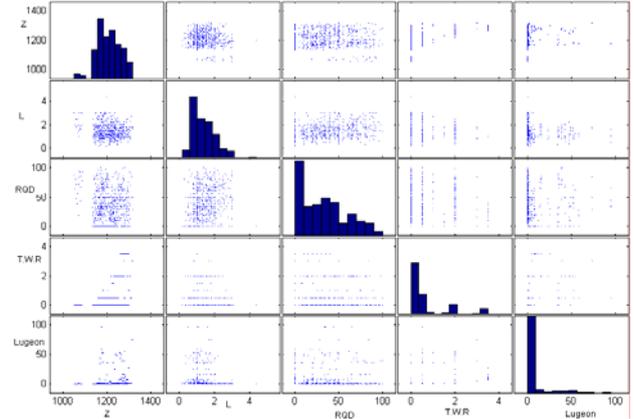

Fig.8. Real data set-Z,L,RQD,T.W.R&lugeon- in matrix plot form (as training data set to SONFIS&SORST )

Table1. The reveled codes of Type of Weathering Rock (TWR), MW: Medium Weathering, SW: Slightly Weathering, CW: Clay Weathering, HW: High Weathering;

| Type of weathering | Ascribed code |
|---|---|
| Fresh-MW | 1.5 |
| SW-MW | 2 |
| Fresh-SW | .5 |
| Fresh | 0 |
| MW | 3 |
| CW | 2.5 |
| SW | 1 |
| HW-MW | 3.5 |
| HW | 4 |

It must be noticed that for unrecognizable objects in test data (elicited by rules) a fix value such 4 is ascribed. So for measure part when any object is not identified, 1 is attributed. This is main reason of such swing of MSE in reduced data set 6 (figure 15-b). Clearly, in data set 7 SORST gains a lowest error (26 neurons in SOM). The extruded rules in the optimum case can be purchased in table 2. We have explained application of SORST in back analysis in other study [14].

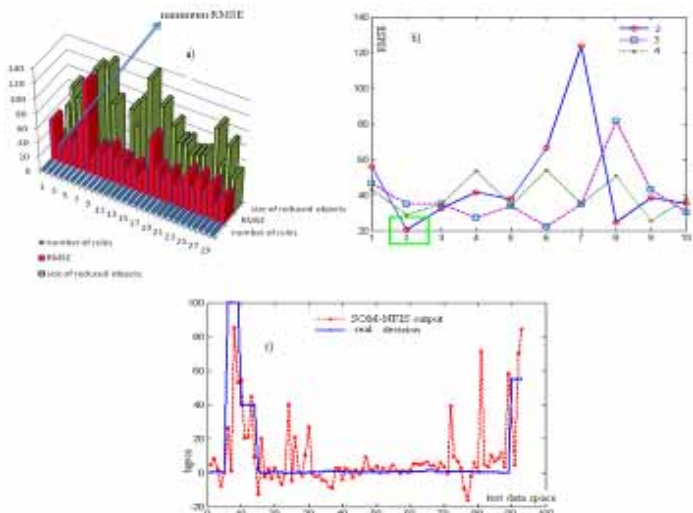

Fig.9. (a&b) SONFIS-R results with maximum number of rules is 4 and close-open iterations is 10; c) Answer of selected SONFIS-R on the test data

Now, we investigate direct application of RST and NFIS on the local coordinates of dam site (as conditional attributes) and lugeon values (as decision part) to depict 3D Iso-surfaces of lugeon variations diagrams.

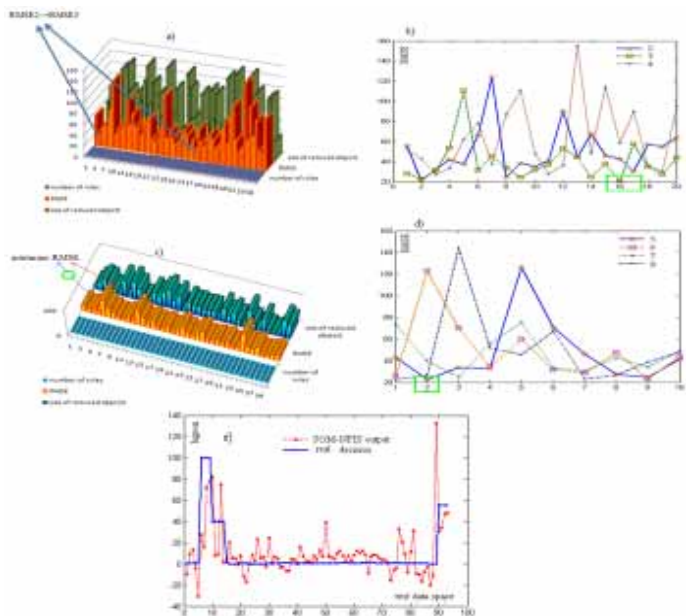

Fig.10. (a &b) SONFIS-R results with maximum number of rules 4 and close-open iterations 20;(c ,d SONFIS-R with 5 to 8 rules number variation and 10 close-open iterations &e) Answer of selected SONFIS-R based on *n.r=5,* on the test data

Figure 17 shows the variation of the lugeon data in Z*= {1} to {5} which has been acquired by serving five condition attributes in RST (figure 16; the symbolic values by 1-D SOM -5 neurons). The categories 1 to 5 state: very low, low, medium, high, and very high, respectively. Number 6 (more than 5) characterizes ambiguity and unknown cases.

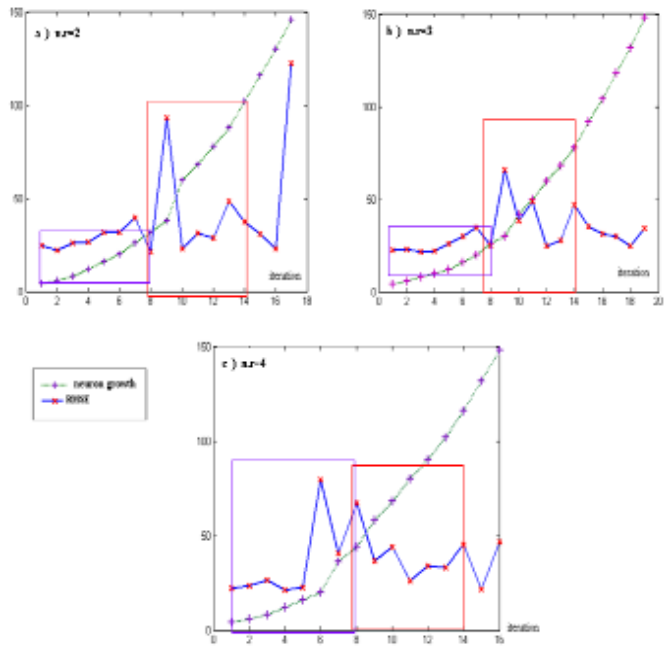

Fig.11.SONFIS-AR: neuron growth & error fluctuations vs. iteration; $\alpha = 1.01$  a) number of rules (*n.r*) =2; b) *n.r*=3 and c) *n.r*=4

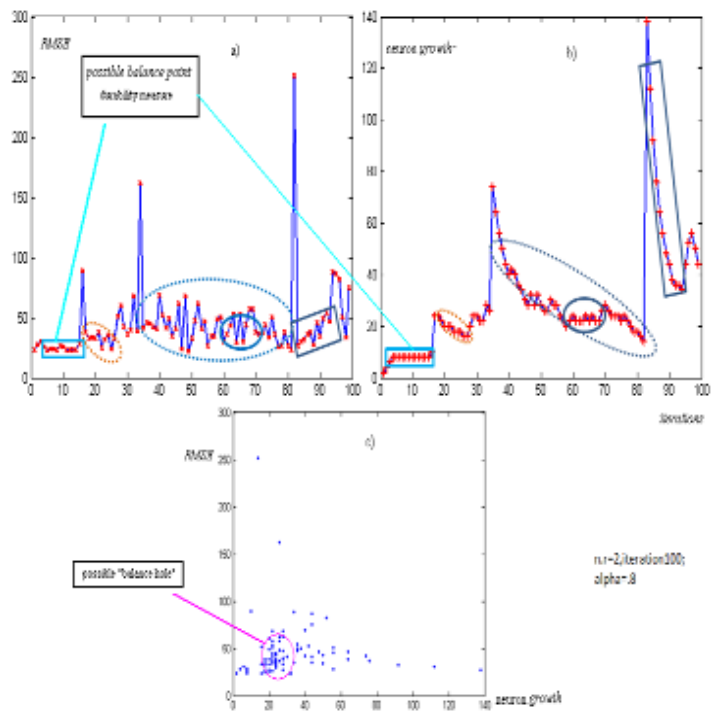

Fig.12. SONFIS-AR: neuron growth & error fluctuations vs. iteration; $\alpha = .8$ - number of rules =2-a) RMSE-iteration; b) neuron growth-iteration c) RMSE- neuron fluctuation: congestion of points can be used as a "balance hole"

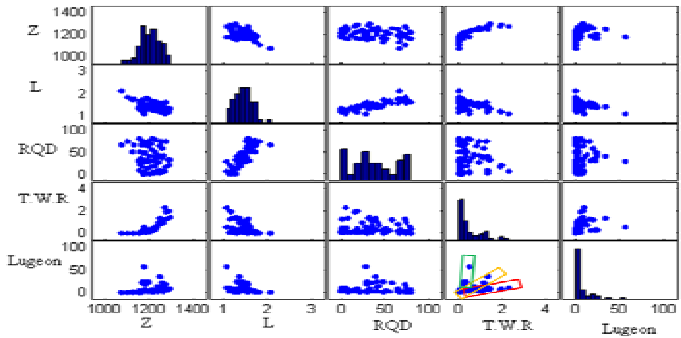

Fig13. Matrix plot of crisp granules by 7*9 grid topology SOM after 500 epochs on the training data set

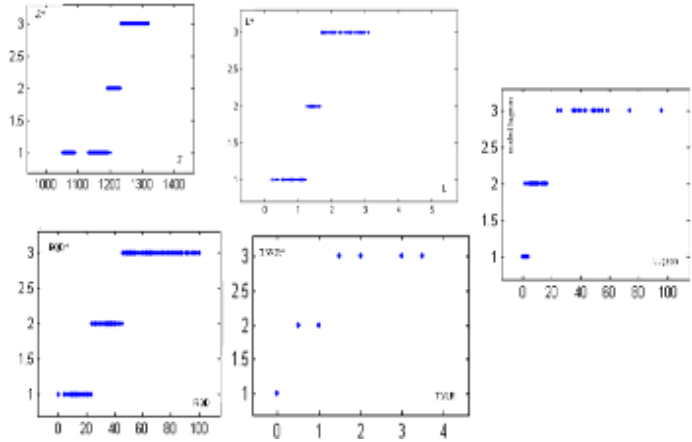

Fig. 14. Results of transferring attribute (Z, L, RQD, T.W.R, and lugeon) in three categories (vertical axes) by 1-D SOM

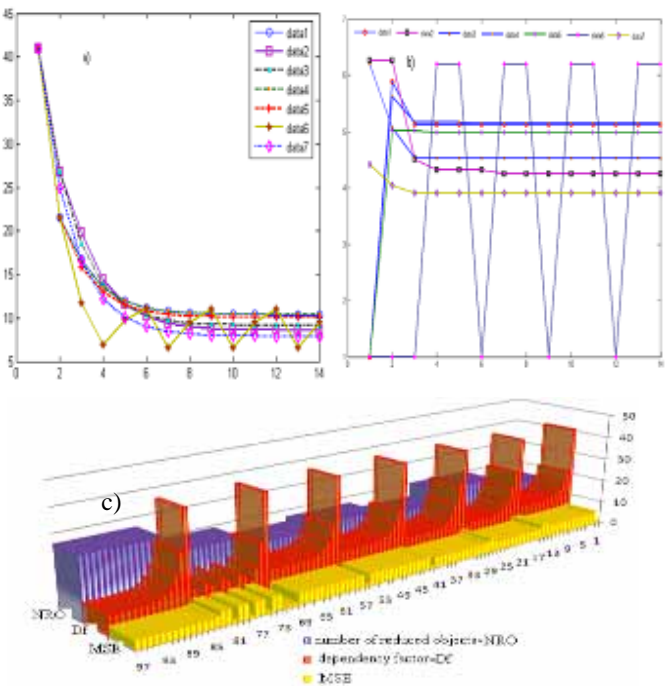

Fig.15. SORST-R results on the lugeon data set: a) strength factor; b) error measure variations along strength factor updating and c) 3-D column perspective of error measure-neuron changes

Table2. Rules on N=26 selected among 696 objects; by SORST-R

| 1 | (z = 2) => (Dec = 1); |
|---|---|
| 2 | (l in {2, 3}) & (rqd = 2) => (Dec = 1); |
| 3 | (z = 3) & (l = 2) & (rqd = 1) => (Dec = 3); |
| 4 | (l = 2) & (twr = 3) => (Dec = 3); |
| 5 | (z = 3) & (l = 1) => (Dec = 1) OR (Dec = 3); |
| 6 | (l in {1, 2}) & (twr = 2) => (Dec = 2); |
| 7 | (rqd = 2) & (twr = 3) => (Dec = 2) OR (Dec = 3); |
| 8 | (z = 1) & (rqd = 1) => (Dec = 2); |

To clarify of permeability changes, in consequent part of rules, the lower value on the symbolic lugeon values which have relatively similar category, for example 1,2,3 or 2,3 or 3,4,5, have been considered. With serving NFIS on such attributes(X, Y, Z& lugeon- without scaling), permeability variations in figures 18 has been portrayed.

In this step, three MFs (Gaussian just as like SONFIS) for input parameters have been utilized. In Consequent of comparison between the results of RST and NFIS, one may interprets the variations in z= {2} is the superposition of sub levels, involved z=1160 to 1200 by NFIS, approximately. So, the compatibility of the results, derived from RST and NFIS can be probed by comparison of figurs17&18. The forecasted domains-dark colors- in figure17, by RST, have been coincided by same regions in figure 18, closely. It must be noticed that the RST model hasn't covered the high permeability zones, because of employing conservative way in estimation of decision part whereas the NFIS has exposed such possible territories. The rate of lugeon variations, or density of permeable parts, distinguishes the zones with capability of possible spring or hole. Such cavities in the dam structures discussed as "karsts", which are the main characteristics of the limestone deposits (figure 20).

To find out the correlation between effective parameters and procuring of valid patterns of the rock mass- in the dam site- one may employ the similar process of NFIS or RST to estimate alterations of RQD and T.W.R (figure 19 using 3 MFs in NFIS). The contrary outputs in some zones with general contextual associated rules about RQD and lugeon, implicate to the relatively complex structures aboard the rock mass.

Apart from a few details, comparison of results indicates three overall zones in the rock mass: in first zone the theoretic rules (such reverse relate between RQD& lugeon) are satisfied, but in other zones, the said rule is disregarded.

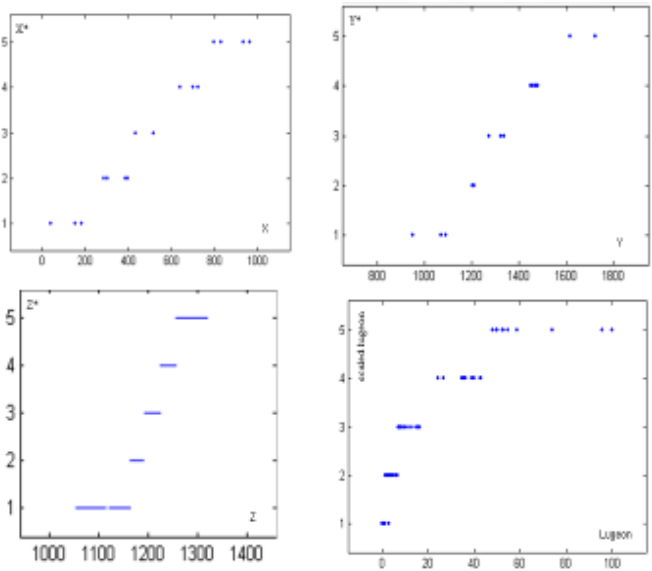

Fig 16. Results of transferring attributes(X, Y, Z and lugeon) in five categories by 1-D SOM

To finding out of the background on these major zones, we refer to the clustered data set by 2D SOM with 7*9 weights in competitive layer (figure 10-c), on the first set of the attributes. The clustered and graphical estimation disclose suitable coordination, relatively. For example in figure 13-b, we have highlighted three distinctive patterns among lugeon and Z, RQD, TWR. One of the main reasons of being such patterns in the investigated rock mass is in the definition of RQD. In measurement of RQD, the direction of joints has not been considered, so that the rock masses with appropriate joints may follow high RQD.

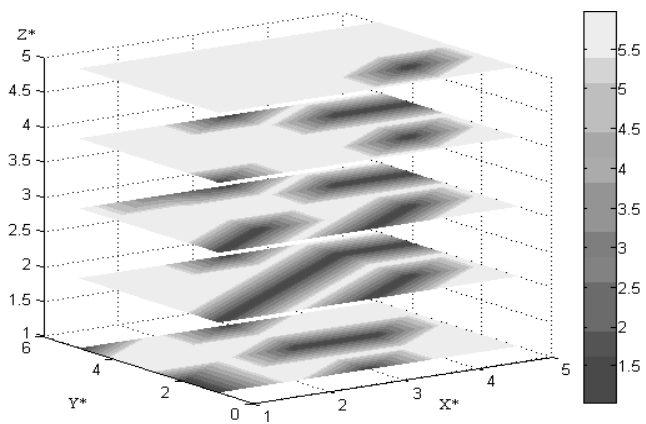

Fig 17. Lugeon variations graphs in z= {1} to z= {5}; accomplished by RST and five scaling of attributes. Number 6 characterizes ambiguity and unknown cases

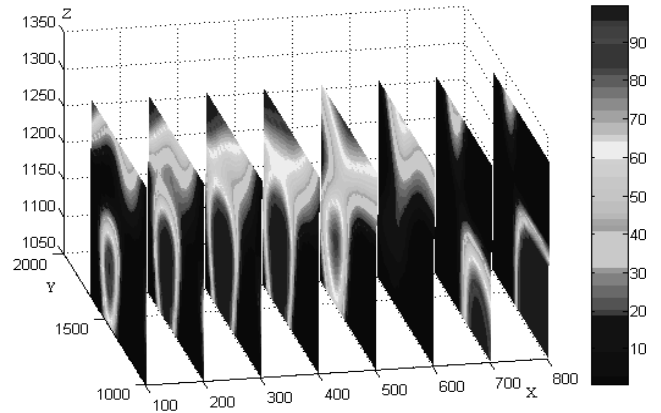

Fig. 18. A cross section perspective of lugeon changes obtained by NFIS

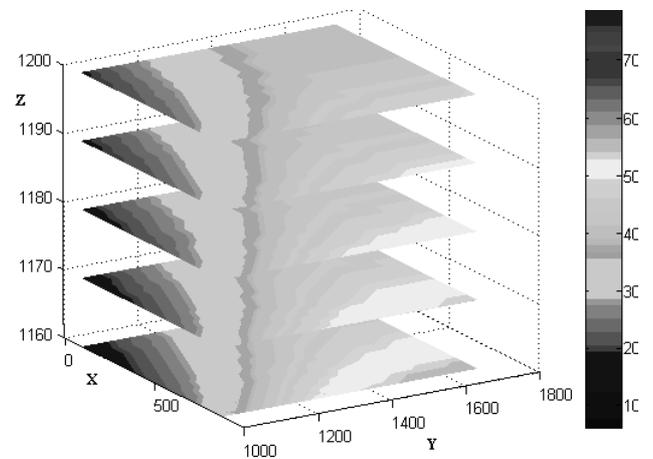

Fig .19. RQD variations in Z= 1160 to Z= 1200; by NFIS

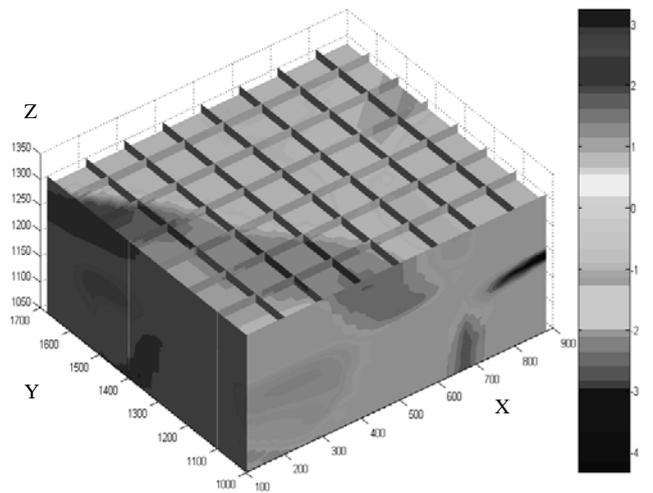

Fig.20. The rate of lugeon variations-possible springs (negative) and cavities (positive values); on the NFIS predictions (divergence of lugeon values)

5. CONCLUSION

The roles of uncertainty and vague information in geomechnaics analysis are undeniable features.

Indeed, with developing of new approaches in information theory and computational intelligence, as well as, soft computing approaches, it is necessary to consider these approaches to better understand of natural events in rock mass. Under this view and granulation theory, we proposed two main algorithms, to complete soft granules construction in not 1-1 mapping level: Self Organizing Neuro-Fuzzy Inference System (Random and Regular neuron growth-SONFIS-R, SONFIS-AR- and Self Organizing Rough Set Theory (SORST). So, we used our systems to analysis of permeability in a dam site, Iran.

ACKNOWLEDGMENT